\documentclass{article}
\usepackage{ijcai24}

\usepackage{times}
\usepackage{soul}
\usepackage{url}
\usepackage[hidelinks]{hyperref}
\usepackage[utf8]{inputenc}
\usepackage[small]{caption}
\usepackage{graphicx}
\usepackage{amsmath}
\usepackage{amsthm}
\usepackage{amsfonts}
\usepackage{booktabs}
\usepackage{algorithm}
\usepackage{algorithmic}
\usepackage[switch]{lineno}
\usepackage{multirow}
\usepackage{makecell}
\usepackage{caption}

\title{EventLens: Leveraging Event-Aware Pretraining and Cross-modal Linking Enhances Visual Commonsense Reasoning}

\author{
	Mingjie Ma$^1$\and
	Zhihuan Yu$^1$\and
	Yichao Ma$^1$\And
	Guohui Li$^2$\thanks{Corresponding Author}\\
	\affiliations
	$^1$School of Computer Science and Technology, Huazhong University of Science and Technology, China\\
	$^2$School of Software Engineering, Huazhong University of Science and Technology, China\\
	\emails
	\{mamj1031, zhihuanyu, myc, guohuili\}@hust.edu.cn
}

\begin{document}

\maketitle

\begin{abstract}

 Visual Commonsense Reasoning (VCR) is a cognitive task, challenging models to answer visual questions requiring human commonsense, and to provide rationales explaining why the answers are correct. With emergence of Large Language Models (LLMs), it is natural and imperative to explore their applicability to VCR. 
 However, VCR task demands more external knowledge to tackle its challenging questions, necessitating special designs to activate LLMs' commonsense reasoning abilities. 
 Also, most existing Multimodal LLMs adopted an abstraction of entire input image, which makes it difficult to comprehend VCR's unique co-reference tags between image regions and text, posing challenges for fine-grained alignment. 
 To address these issues, we propose EventLens that leverages \textbf{Event}-Aware Pretraining and Cross-modal \textbf{L}inking and \textbf{En}hance\textbf{S} VCR.
 First, by emulating the cognitive process of human reasoning, an Event-Aware Pretraining auxiliary task is introduced to better activate LLM's global comprehension of intricate scenarios. 
 Second, during fine-tuning, we further utilize reference tags to bridge RoI features with texts, while preserving both modality semantics.
 Finally, we use instruct-style prompts to narrow the gap between pretraining and fine-tuning, and task-specific adapters to better integrate LLM's inherent knowledge with new commonsense. 
 Experimental results show the effectiveness of our proposed auxiliary task and fine-grained linking strategy.
   
\end{abstract}

\section{Introduction}

The field of multimodal interaction involving visual and language modalities has recently witnessed significant attention, with numerous noteworthy tasks such as cross-modal retrieval~\cite{retrieval}, image captioning\cite{coco:chen,concap} and visual question answering (VQA)~\cite{vqa}. However, researchers \cite{r2c:zeller} found that typical multimodal reasoning tasks, like VQA, mainly focused on trivia recognition questions (e.g., \textit{how many}, \textit{what color} or \textit{is something present}), and felt the need to endow models with cognitive-level capabilities so that Visual Commonsense Reasoning (VCR) task was formulated.

\begin{figure}[t]
	\centering
	\includegraphics[width=0.9\columnwidth]{{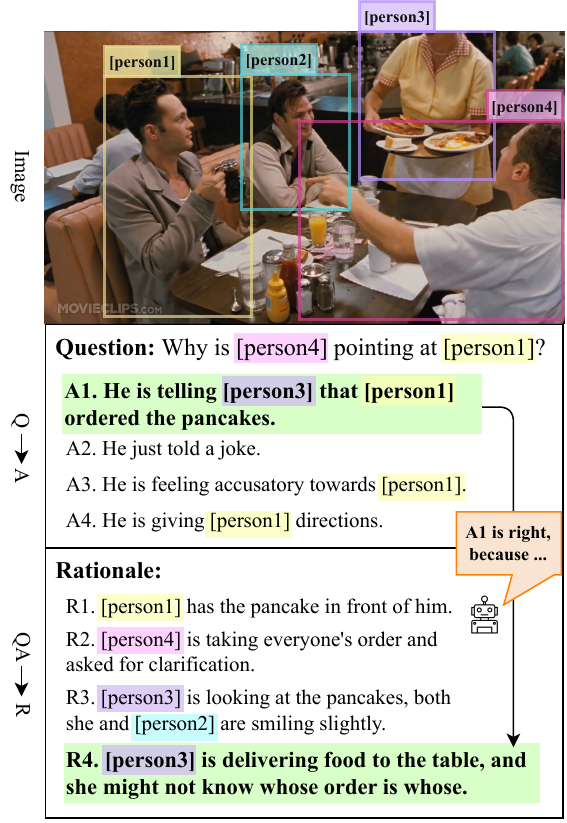}}
	\caption{An example of VCR with two subtasks.}
	\label{fig1}
\end{figure}

Figure \ref{fig1} shows a typical VCR example. When presented with a cognitive question and a complex image, models are first required to deduce the right answer and then its right rationale, namely question-answering (Q$\rightarrow$A) and answer-justification (QA$\rightarrow$R) subtasks. Compared to traditional VQA task, VCR is more challenging as it often contains counterfactual or dynamic questions, which compels models to go beyond mere fitting of dataset distributions. Instead, models must genuinely possess commonsense comprehending and reasoning abilities to demonstrate superior performance when conquering those two subtasks.

To address the VCR task, models tailored specifically for this challenge are proposed, such as ECMR~\cite{ecmr:zhang} and CCN~\cite{ccn:wu}. These methods typically employ carefully designed attention mechanisms~\cite{r2c:zeller,tab:lin}, graph structures~\cite{sgeitl:wang,hgl:yu}, and/or other methods~\cite{ccn:wu,multicl:zhang} to align more closely with human reasoning intuition. On the other hand, there is a strand of research adopting methods based on Vision Language Transformers (VLT)~\cite{attn:vaswani,vltemp:li}. They usually start with a pretrained transformer~\cite{uniter:chen,kvl-bert:song}, and then fine-tune it by jointly inputting text embeddings and visual region features on VCR task. These VLT-based models have shown significant performance improvements in the VCR task.

With the emergence of large language models, like GPT~\cite{gpt:brown}, OPT~\cite{opt:zhang} and LLaMA~\cite{llama}, in the past two years, an increasing number of studies~\cite{bliva,llava,blip2:li}  have begun exploring how to leverage these pretrained language models for multimodal tasks. These approaches have achieved notable success in recognition tasks (such as VQA). A natural question arises: can these methods be further utilized for cognitive tasks? By incorporating large language models into the VCR task, we can leverage their outstanding capabilities in commonsense knowledge and language understanding to handle the fusion of visual and textual information more comprehensively and holistically. The advantage of using LLMs lies in the ability to better understand and interpret image content through learning context and reasoning abilities, thereby demonstrating superior performance in tasks involving answering questions and providing rationales.
%

However, compared to recognition tasks, the cognitive task of VCR poses some key challenges while incorporating with LLMs:

\begin{enumerate}
	\item In the VCR task, visual scenes are often more intricate, and the questions more challenging. Existing task-specific models lack sufficient knowledge for effective commonsense reasoning. While pretrained models have acquired rich human commonsense knowledge paradigms from large scale of training data, adapting them to VCR requires a significant cost in fine-tuning.
	\item The VCR dataset itself has certain limitation that, instead of using natural language reference to identify a visual object in the image, it directly uses bounding box tags of visual object regions to reference them. This design makes it challenging for visual language models with structures similar to BLIP-2 to align text tokens with objects in the image. Addressing this challenge may require further efforts, to leverage the annotated object tag information to better support VCR task.
\end{enumerate}

To address the aforementioned challenges, we propose EventLens, a Multimodal LLM architecture that leverages \textbf{Event}-Aware Pretraining and Cross-modal \textbf{L}inking and \textbf{En}hance\textbf{S} VCR.
Specifically, we first introduce an Event-Aware Pretraining task, aiming to enable the model to understand complex visual scenes while focusing on objects in the image and to infer ongoing events and the intentions of characters within the event scene, thereby enhancing the model's commonsense knowledge reasoning abilities. Secondly, we introduce a fine-grained Cross-modal Local Linking method. By freezing the visual backbone and Q-Former, we independently process object regions and then fuse them with corresponding text objects. This strategy aims to strengthen the model's understanding of the correlations between text tokens, local object features, and holistic image scene, thus improving performance in the VCR task. Finally, we use instruct-style prompts to narrow the gap between pretraining and VCR tasks, and task-specific adapters to better integrate LLM's inherent knowledge with new data paradigm. These approaches not only help acquire commonsense knowledge for sophisticated scenario but also avoid from significant cost of full fine-tuning multimodal LLM.

Experiments on VCR dataset demonstrate outstanding performance of EventLens, showing that it achieves competitive performance with a relatively small number of trainable parameters which reduces the computational resource costs of training. Ablation experiments further prove the effectiveness of EventLens. Our work provides a comprehensive and forward-looking approach to enhance the performance of deep learning models in multimodal reasoning tasks.

\section{Related Works}
In this section, we review related work on Visual Commonsense Reasoning and Multimodal Large Language Models.

\subsection{Visual Commonsense Reasoning} 
VCR is proposed as a multimodal cognitive task~\cite{r2c:zeller}. The objective of VCR task is to infer answers to questions based on images and provide justifications for the correct answers. 
Researchers have employed various approaches to address these challenges. 
Some have focused on developing task-specific models. 
For instance, R2C~\cite{r2c:zeller} utilizes holistic attention mechanism and LSTM to contextualize questions, responses and image objects for reasoning, aiming to approach cognitive-level understanding.
TAB-VCR~\cite{tab:lin} associates visual features with attribute information and enhances grounding of noun phrases through additional object detection.
HGL~\cite{hgl:yu} proposes a heterogeneous graph learning architecture to bridge semantic gap between visual and language domains.
SGEITL\cite{sgeitl:wang} uses a multi-hop graph transformer and scene-aware pretraining method to leverages rich structural information from scene graphs.
Inspired by neuroscience, CCN~\cite{ccn:wu} introduces a Connective Cognition Network, which dynamically reorganizes contextualized visual neuron connectivity based on semantics of texts.
Zhang et al.~\shortcite{multicl:zhang} introduces a multi-level counterfactual contrastive learning framework to jointly model hierarchical visual content and the modal relationship between vision and language

Additionally, recognizing that the VCR dataset may not sufficiently enable models to acquire human commonsense knowledge, efforts have been made to introduce external knowledge. \cite{multiki:wen} proposes a Knowledge Transfer Network, employing a multi-level knowledge transfer approach to extract knowledge from commonsense reasoning datasets to help form reasoning cues for VCR.

Other than task-specific approaches, researchers have also leveraged pretrained Vision Language Transformers (VLT)~\cite{vlpsurvey:du,vl-bert,ernie-vil:yu}, which have demonstrated exceptional performance in vision-language tasks. Recently VLT-based methods have emerged leading in VCR task~\cite{uniter:chen,vilbert:lu,ernie-vil:yu,pevl:yao}. A recent study~\cite{vltemp:li} suggests that existing VLT-based models have not fully exploited the unique tag labels present in VCR. B2T2~\cite{b2t2:alberti} also conducted empirical research on VLT, indicating that global features and a sufficient number of region-of-interests (ROIs) can effectively enhance VCR task performance. Inspired by these studies, we propose leveraging LLMs for its wide range of knowledge source,while simultaneously bridging visual region features and text modality making use of inherent tag labels to enhance understanding of challenging multimodal scenes in VCR.

\subsection{Multimodal Large Language Models}

Large Language Models (LLMs) have demonstrated impressive zero- and few-shot capabilities across various NLP tasks and applications. However, their adoption in other modalities poses challenges. Recent research has explored leveraging pretrained LLMs to construct multimodal unified models. For instance, Flamingo~\cite{flamingo} connects a visual encoder and LLM through a perceiver resampler, demonstrating outstanding few-shot performance. BLIP-2~\cite{blip2:li} designs Q-Former to align visual features with LLM. MiniGPT-4~\cite{minigpt} employs the same Q-Former but replaces LLM with Vicuna~\cite{vicuna}. LLaVA~\cite{llava} directly fine-tunes LLMs and mPlug-OWL~\cite{mplug} efficiently tunes LLaMA through LoRA adaption~\cite{lora}.

Despite their excellent performance on image-text tasks, these methods often struggle to focus exclusively on local details in images due to the use of abstract modules with learned query embeddings, such as Q-Former. Recognizing this limitation, BLIVA~\cite{bliva} combines Q-Former output and complete image features as input to LLM, allowing LLM to capture both abstract and specific visual information. However, introducing complete image features may incur significant computational overhead and introduce noise for VCR task. Therefore, our work places a greater emphasis on incorporating abstract queries from region-of-interests (ROIs) to address this concern.

\section{Methodology}

We first present the problem formulation, and then briefly introduce our observation and intuition, based on which we propose our model architectures as well as training stages.

\subsection{Preliminaries}

\subsubsection{Problem Formulation}

As illustrated in Figure \ref{fig1}, a VCR task sample can be represented as a quintuple $[I, O, Q, A, R]$, where $I$ is an image, $O$ includes a set of bounding box coordinates indicating the regions of interest with reference tags (such as $[person1]$) in the image, $Q$ denotes a question about the image, $A$ consists of four answer options $\{A_i, i=1, ...\ ,4\}$ and $R$ four rationale options. The question-answering subtask requires predicting the correct answer $A_g$ from $A$, while answer-justification subtask choosing the correct rationale $R_g$ from $R$, which best supports $A_g$ as the correct answer to the question $Q$ (subscription $g$ means $ground\ truth$).

\subsubsection{Intuitive Observations}
We make some intuitive yet meaningful observations about incorporating LLMs with VCR: 
\begin{enumerate}
	\item Firstly, to reason over a question based on an image, it is helpful to know in advance where you are and what is happening in the surroundings.
	\item Secondly, even if previous research on multimodal LLMs such as BLIP-2 and InstructBLIP demonstrates the ability to generate diverse and detailed descriptions for images, they still cannot determine which specific individual is being referred to in the VCR tests, probably because query tokens of Q-Former act as an abstract when the reference (such as [person1]) lacks textual contexts.
\end{enumerate}

\subsection{Event-Aware Pretraining}
 Based on our first observation, before humans answer more challenging questions in a complex environment, we first observe and understand the environment, determine their location, deduce relationships between visual objects in the environment based on the question, and then we answer the question.

\begin{figure}[t]
	\centering
	\includegraphics[width=\columnwidth]{{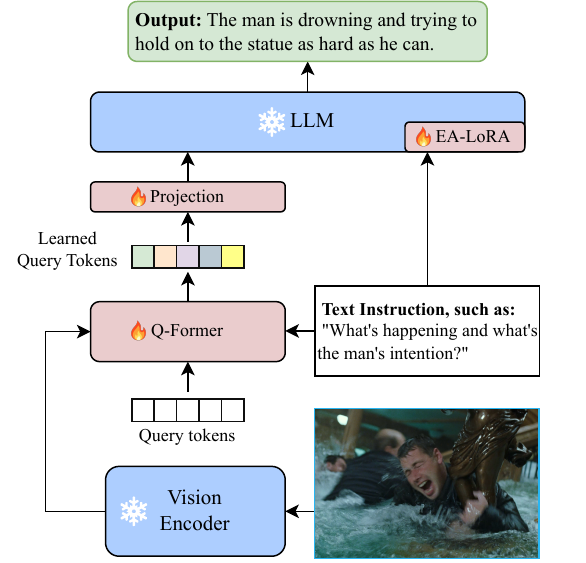}}
	\caption{EventLens Architecture for Event-Aware Pretraining.}
	\label{fig2}
\end{figure}

Drawing an analogy to human cognitive processes, our objective is to facilitate the model's adaptation and endow it with the capacity to describe complex visual environments before engaging in the actual VCR task. To achieve this, we propose a new pretraining stage called Event-Aware Pretraining, as illustrated in Figure \ref{fig2}. The primary aim of this stage is to empower Q-Former with the proficiency to convey intricate visual cues in challenging datasets like VCR, effectively unleashing the potential of language model.

We adopt and make some modification to BLIP-2 architecture~\cite{blip2:li}. Specifically, an input image is first patchified and processed through a Vision Transformer (ViT) as vision encoder with frozen parameters for visual features:

\begin{equation}
	H_v = ViT(f_p(I)) \in \mathbb{R}^{p \times d_v},
\end{equation}
where $f_p(\cdot)$ splits an image into $p$ non-overlapping patches, $ViT(\cdot)$ stands for vision encoder and $d_v$ is the dimension of encoded patch embedding.

Then a trainable module, Q-Former, takes in a learnable query sequence $H_q \in \mathbb{R}^{N \times d_q}$ concatenated with our predefined instruction, to extract a fixed number of output learned queries from $H_v$, where $d_q$ is the hidden states' dimension of Q-Former. 

We intent to instruct the model to generate dynamic descriptions of complex scenes before answering related questions. This is achieved by constructing instructions like ``\textit{Describe the image}", ``\textit{What happened before?}" or ``\textit{What is going to happen in this scene?}", noting that the instructions demand our model to pre-reason over an intricate visual scene.
The instructions are tokenized and embedded into word embeddings $T_{ins}=[t_m]_{m=1}^M$, where $M$ is the number of instruction tokens. We concatenate $H_q$ and $T_{ins}$, and then feed the combination into Q-Former to extract \textit{instruction-related} query tokens from image features $H_v$. 

Next, the learned queries will be projected into the downstream language model's latent space:

\begin{equation}
	H_q' = Proj_l(f_{QF}([H_q, T_{ins}], H_v)) \in \mathbb{R} ^ {N \times d_l},
\end{equation}
where we define the Q-Former operation as a function $f_{QF}(\cdot)$, $H_q'$ is the projected learned query tokens, and $d_l$ is dimension of LLM latent space.

The instruction embeddings will then again be concatenated with $H_q'$ and input to LLM, as $T=[H_q', T_{ins}]=[h_i]_{i=1}^{N+M}$, to generate instruction-following texts. As shown in Figure \ref{fig2}, the model is instructed to describe the man's intention in such a dynamic scene, and the model correctly follows the instruction by outputting ``\textit{the man is drowning and trying to hold on to the statue...}". 

In addition, since we want to avoid huge training overhead and catastrophic forgetting, while still hoping to enhance adaptability of the language model to such instructions and seamlessly apply its reasoning and generation ability to understand and describe dynamic processes in static frames, we choose to inject a LoRA-based adapter into LLM during Event-Aware pretraining, namely \textit{EA-LoRA}.

Specifically, during the self-attention process in LLM, the LoRA-based adapter is incorporated into the projection layers of Query and Value (the \textit{Query} here is different from the learnable \textit{Query} tokens in Q-Former operation):

\begin{equation}
	\begin{aligned}
		Query & = W_q T + \Delta Q , \\
		Key & = W_k T , \\
		Value & = W_v T + \Delta V , 
	\end{aligned}
\end{equation}

\begin{equation}
	\begin{aligned}
		\Delta Q & = W_q^{up} (W_q^{down} T) ,\\
		\Delta V & = W_v^{up}  (W_v^{down} T) ,
	\end{aligned}
\end{equation}
where $W_q$, $W_k$ and $W_v$ are frozen LLM self-attention projection matrix, $W_*^{up} \in \mathbb{R}^{d_a \times d_l}$ and $W_*^{up} \in \mathbb{R}^{d_l \times d_r}$, ($* \in \{q, v\}$) are down- and up-projection matrix in the adapter, and $d_r$ is a hyperparameter of EA-LoRA. During this stage, only Q-Former, EA-adapter and projection layer are trainable, where the number of EA-LoRA parameter is less than 0.5\% that of LLM. Since Event-Aware pretraining is still a language generation task, we pretrain this stage with language modeling loss.


\subsection{Supervised Fine-tuning for VCR}

\begin{figure*}[t]
	\centering
	\includegraphics[width=1.8\columnwidth]{{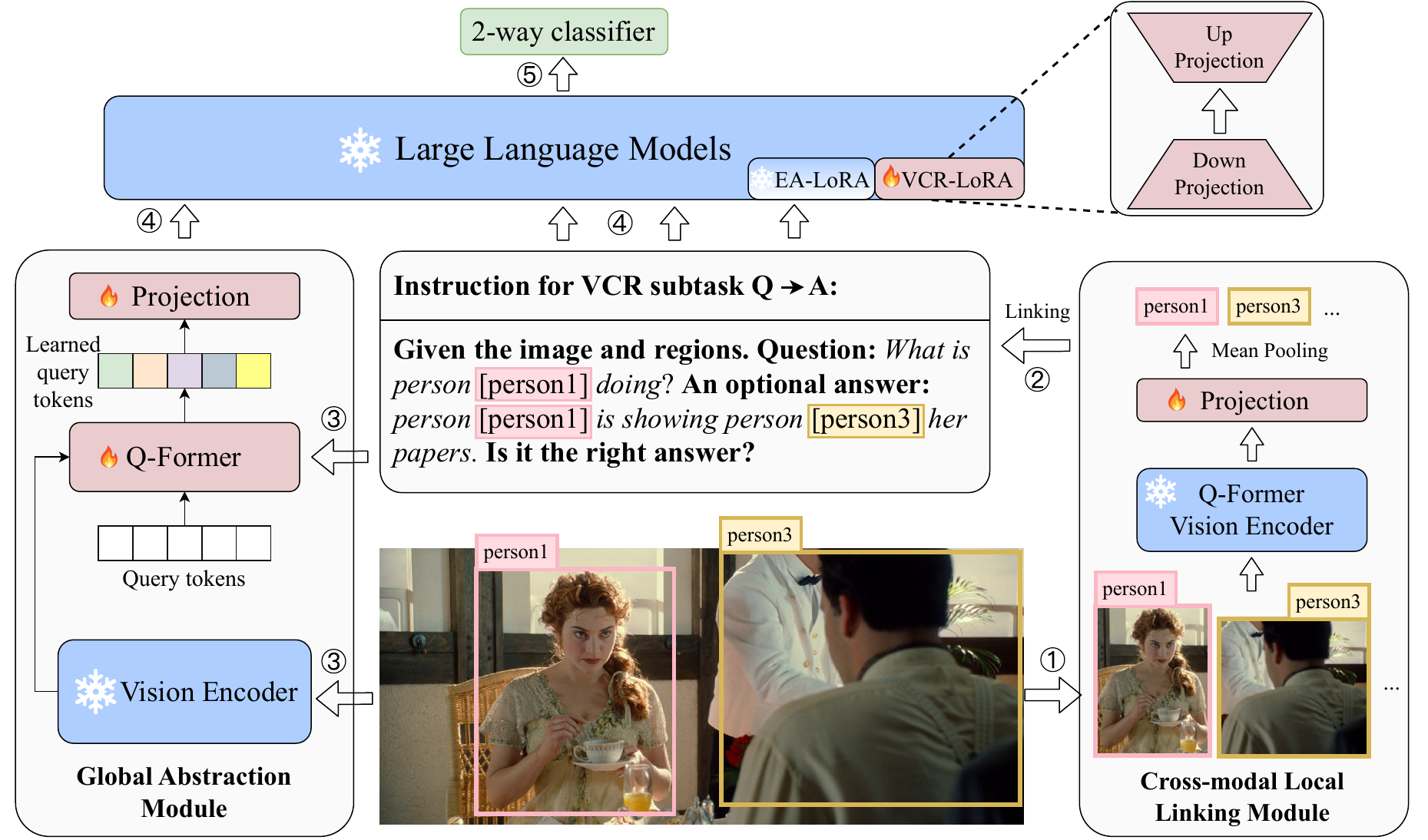}}
	\caption{Proposed EventLens Architecture, including (1) a Global Abstraction module to extract instruction-related image features; (2) a Cross-modal Local Linking module, to better solve the VCR co-reference lacking problem and better instruct LLM, and (3) downstream reasoning core LLM to predict instruction-following answers. The numbered marks denote the steps of EventLens architecture workflow.}
	\label{fig3}
\end{figure*}

During Supervised Fine-tuning stage, we adopt the architecture depicted in Figure~\ref{fig3}, which contains three major components: a Global Abstraction module, a Cross-modal Local Linking module and a downstream reasoning core LLM.

\subsubsection{Global Abstraction module}
The Global Abstraction module on the left part of Figure~\ref{fig3} is similar to previous Event-Aware pretraining stage with slightly different input instructions. The instructions for VCR fine-tuning are thus formed:
\begin{itemize}
	\item For Question Answering (Q$\rightarrow$A) subtask, we fill in the question $Q$ and answer option $A_i$ into the predefined instruction template. Then the model can be instructed to decide whether it is a right answer to the question, as shown in Figure~\ref{fig3}.
	\item For Answer Justification (QA$\rightarrow$R) subtask, similarly, the question $Q$, correct answer $A_g$ and a rationale option $R_i$ will be filled into the instruction template to instruct our model to decide whether the rationale best justify answer $A_g$ to question $Q$.
\end{itemize}
Additionally, the holistic setting (Q$\rightarrow$AR) requires the model to correctly finish both the (Q$\rightarrow$A) and (QA$\rightarrow$R) subtasks.

\subsubsection{Cross-modal Local Linking module}
According to our second observation, when it comes to a specific visual object, the VCR's text (including question, candidate answers and rationales) refers to the object in the image by directly using its bounding box tag. For example, the question in Figure 3 is ``\textit{what is [person1] doing}", using a tag [person1] to refer to ``\textit{the woman holding her cup}", but such question context does not contain enough information to indicate which object is being asked about. 

This approach may seem intuitive in human communication, resembling pointing to an object in a scene with one's finger while asking a question. However, for recently popular multi-modal LLMs based on Q-Former, without description providing co-reference information, they can only ``guess" whichever object the tag \textit{[person1]} refers to.

In order to address this issue, we introduce a \textbf{C}ross-modal \textbf{L}ocal \textbf{L}inking (\textbf{CLL}) module, which, while preserving textual semantics and visual features, connects corresponding objects between the visual and textual domains based on reference tags.

Specifically, for a VCR sample, our CLL module begins by extracting reference tags from the input text (i.e., $Q$, $\{A_i\}$ and $\{R_i\}$), which are used to query corresponding objects' bounding box coordinates in the object set $O$. Then we use these coordinates to crop out a set of object sub-images $I_{sub} = \{o_i\}_{i=1}^K$ from input image $I$, where each object sub-image $o_i$ corresponds to a reference tag, and $K$ is the number of objects cropped from $I$.

Subsequently, each sub-image undergoes processing by a sequence of frozen networks: a visual encoder, a Q-Former pretrained in the previous Event-Aware stage, and a trainable projection layer. This results in a fix number of representations for each visual object, embedded into the latent space of downstream language model. Utilizing average pooling, we obtain a single-token representation for each object. 

Finally, by concatenating the text word embeddings from the tags with the corresponding visual feature tokens and replacing the original reference tags, we establish links between objects that should be aligned between the visual and text domains. Take a reference tag [person1] as an example:

\begin{equation}
	f_{CLL}(\text{[person1]}) = [ f_{word}(\text{``person"}), f_{sub}(o_1)],
\end{equation}
where $f_{CLL}(\cdot)$ is the proposed Cross-model Local Linking module process, $f_{word}(\cdot)$ is the word embedding layer and $f_{sub}(\cdot)$ is the sub-image's single token extraction process just illustrated. Our $f_{CLL}(\cdot)$ module receives the reference tag \textit{[person1]}, and maps it to a two-token combination, preserving the semantic information of the tags (e.g., "person") and achieving explicit cross-modal linking between textual references and visual region features. In this way, the CLL module explicitly utilizes the object tags in the VCR dataset and instruct LLM clearly.

\subsubsection{Reasoning Core LLM}
Considering that the text style in the VCR task is very different from pretraining tasks, we freeze the EA-LoRA  in Event-Aware pretraining stage, and introduce a new LoRA-based VCR adapter (noted as VCR-LoRA in Figure~\ref{fig3}) identical to EA-LoRA. Finally, we use a binary classification head to predict whether the answer/rationale option is correct based on the LLM's output next token with Cross Entropy Loss.

\section{Experiments}

 In this section, we introduce the datasets and implementation details of our approach. Then we report the main quantitative results, and compare with recent state-of-the-arts. Finally ablation studies are conducted to investigate the contribution of training methods and proposed components of our work.

\subsection{Datasets}
\begin{table}[t]
	\centering
	\resizebox{\columnwidth}{!}{%
		\begin{tabular}{c|l}
			\hline
			\textbf{Task} &
			\multicolumn{1}{c}{\textbf{Instruction Templates}} \\ \hline \hline
			\multirow{4}{*}{\textbf{\begin{tabular}[c]{@{}c@{}}Event-Aware\\ Pretraining\end{tabular}}} &
			(1) Describe the image in detail. \\
			&
			(2) What's happening and what's the person's intention? \\
			&
			(3) What was happening before this scene? \\
			&
			(4) What will happen later? \\ \hline \hline
			\multirow{3}{*}{\textbf{\begin{tabular}[c]{@{}c@{}}Q-A\\ subtask\end{tabular}}} &
			\begin{tabular}[c]{@{}l@{}}(1) Given the image and regions. Question: $Q$ \\ An optional answer: $A_i$   Is it the right answer?\end{tabular} \\
			&
			\begin{tabular}[c]{@{}l@{}}(2) Based on the image. Does $A_i$ best answer \\ the question $Q$?\end{tabular} \\
			&
			\begin{tabular}[c]{@{}l@{}}(3) Take into account scene information in the image. \\ Question: $Q$ An answer choice: $A_i$ Is the answer \\choice right?\end{tabular} \\ \hline \hline
			\multirow{3}{*}{\textbf{\begin{tabular}[c]{@{}c@{}}QA-R\\ subtask\end{tabular}}} &
			\begin{tabular}[c]{@{}l@{}}(1) Given the image and regions. Question: $Q$ \\ The right answer: $A_g$ An optional rationale: $R_i$\\ Is it the right rationale?\end{tabular} \\
			&
			\begin{tabular}[c]{@{}l@{}}(2) Based on the image. Does $R_i$ best justify $A_g$ \\to the question $Q$?\end{tabular} \\
			&
			\begin{tabular}[c]{@{}l@{}}(3) Take into account scene information in the image. \\ Question: $Q$ Right answer: $A_g$ A rationale choice: \\ $R_i$ Is the rationale choice right?\end{tabular} \\ \hline
		\end{tabular}%
	}
	\caption{Instruction templates used for Event-Aware pretraining and two subtasks of VCR fine-tuning}
	\label{tab:ins}
\end{table}

We conducted Event-Aware pretraining on VisualCOMET~\cite{visualcomet}, which comprises a total of 60k images, with 139k Event, 580k Before, 580k After and 295k Intent descriptions. Noting that images in VisualCOMET form a subset of those in VCR, to prevent data leakage, we initially check and make sure there is no duplicate images between the VisualCOMET training set and the VCR validation/test sets. Subsequently, we transform the descriptions in VisualCOMET training set into declarative sentences, generating nearly 1.28M descriptions that require complex reasoning, providing a challenging context for the proposed Event-Aware Pretraining.

Extensive experiments are then carried out on VCR dataset, which consists of 290k multiple-choice questions, derived from 110k movie scenes~\cite{r2c:zeller}. The dataset's images are selected from LSMDC~\cite{lsmdc} and MovieClips\footnote{Fandango MovieClips: \url{youtube.com/user/movieclips}}. We utilized the dataset splits provided by VCR, with training, validation, and test set sizes of 212,923, 26,534, and 25,263 entries, respectively. For Q$\rightarrow$A and QA$\rightarrow$R subtasks, four candidate options are given, with only one being the correct answer. Additionally, VCR includes a Q$\rightarrow$AR setting, where points are received only when both Q$\rightarrow$A and QA$\rightarrow$R are answered correctly. As ground-truth labels for the test set are not publicly available, we conduct most experiments on the validation set, and only report results of our best model on the test set.

\subsection{Implementation Details}

In our experiments, we adopt several instruction templates for both Event-Aware Pretraining and VCR fine-tuning, illustrated in Table~\ref{tab:ins}. Our EventLens architecture incorporates ViT-g/14 from EVA-CLIP~\cite{eva:fang} as vision encoder, and $\text{OPT}_{\text{2.7b}}$ or $\text{OPT}_{\text{6.7b}}$ as downstream LLM, marked as \textbf{EventLens-base} and \textbf{EventLens-large}. We formulated the Event-Aware pretraining task as a Language Modeling task, and VCR as a binary classification task. We use the AdamW~\cite{adamw} optimizer with a learning rate of 1e-5 and a cosine strategy scheduler. The maximum input length for LLM is set to 256. Our EventLens and all the ablated versions are implemented with Pytorch~\cite{torch}, and are trained on a platform with 4 Nvidia RTX4090-24GB GPUs.

\subsection{Main Results}
For a holistic and objective comparison, we compare EventLens versus classic or leading models which can be classified into three groups:
\begin{table*}[t]
	\centering
		\captionsetup{width=0.7\textwidth}
		\begin{tabular}{c|cccccc|c}
			\hline
			\multicolumn{1}{c|}{\multirow{2}{*}{Model}} &
			\multicolumn{2}{c|}{Q-A} &
			\multicolumn{2}{c|}{QA-R} &
			\multicolumn{2}{c|}{Q-AR} &
			\multicolumn{1}{c}{\multirow{2}{*}{\begin{tabular}[c]{@{}c@{}}Trainable \\ Params\end{tabular}}} \\
			\multicolumn{1}{c|}{} &
			\multicolumn{1}{c}{Eval} &
			\multicolumn{1}{c|}{Test} &
			\multicolumn{1}{c}{Eval} &
			\multicolumn{1}{c|}{Test} &
			\multicolumn{1}{c}{Eval} &
			\multicolumn{1}{c|}{Test} &
			\multicolumn{1}{c}{} \\ \hline \hline
			BERT              &53.8  &53.9  &64.1  &64.5  &34.8  &35.0  &--  \\ 
			LSTM+ELMO &28.1  &28.3  &28.7  &28.5  &8.3  &8.4  & -- \\ \hline \hline
			R2C                &63.8  &65.1  &67.2  &67.3  &43.1  &44.0  &--  \\ 
			HGL                &69.4  &70.1  &70.6  &70.8  &49.1  &49.8  &  --\\ 
			CCN               &67.4  &68.5  &70.6  &70.5  &47.7  &48.4  & -- \\
			TAB-VCR       &69.9  &70.4  &72.2  &71.7  &50.6  &50.5  &--  \\
			CKRM            &66.2  &66.9  &68.5  &68.5  &45.6  &45.9  &--  \\
			ECMR             &70.7  &70.4  &72.0  &71.3  &51.1  &50.5  &--  \\  \hline \hline
			UNITER-large &77.2  &77.3  &80.5  &80.8  &62.6  &62.8  &303M  \\
			VL-BERT-large	&75.5 &75.8 &77.9 &78.4 &58.9 &59.7 &340M  \\
			SGEITL   &74.9  &76.0  &77.2  &78.0  &57.8  &59.6  &--  \\
			PEVL   &75.1  &76.0  &76.4  &76.7  &57.8  &58.6  &233M  \\ 
			GPT-4 &--  &73.5  &--  &75.4  &--  &56.2  &--  \\ 
			BLIP-2*  &74.9  &--  &74.2  &--  &56.1  &--  &108M  \\ \hline \hline
			\textbf{EventLens-base} &80.6  &81.0  &79.9  &80.3  &64.7  &65.5  &128M \\ 
			\textbf{EventLens-large}&\textbf{82.0}  &\textbf{82.7}  &\textbf{82.5}  &\textbf{82.7}  &\textbf{67.9}  &\textbf{68.5}  &144M \\ \hline
		\end{tabular}
	\caption{Comparison results on VCR with text-only, task-specific and VLT-based mathods. we also report trainable parameter numbers of open-source VLT-based methods.}
	\label{tab:main}
\end{table*}
\begin{itemize}
	\item Text-only baselines report by~\cite{r2c:zeller}, including \textbf{BERT}~\cite{bert:devlin} and\textbf{ LSTM+ELMO}.
	\item Task-specific models without pretraing, including \textbf{R2C}~\cite{r2c:zeller}, \textbf{HGL}~\cite{hgl:yu},  \textbf{CCN}~\cite{ccn:wu}, \textbf{TAB-VCR}~\cite{tab:lin}, \textbf{CKRM}~\cite{ckrm:wen}, \textbf{ECMR}~\cite{ecmr:zhang}.
	\item VLT-VCR models based on pretraining, including \textbf{UNITER}~\cite{uniter:chen}, \textbf{VL-BERT}~\cite{vl-bert}, \textbf{SGEITL}~\cite{sgeitl:wang}, \textbf{PEVL}~\cite{pevl:yao}, \textbf{GPT-4}, \textbf{BLIP-2}~\cite{blip2:li}.
\end{itemize}

As shown in Table \ref{tab:main}, proposed EventLens significantly outperforms the task-specific models and demonstrates excellent performance on the validation and test sets without a significant increase in the number of trainable parameters compared to existing VLT models. Due to the non-disclosure of labels in the test set, we acquire test set scores by submitting results to the VCR leaderboard. Specifically, compared with the current best graph-based model ECMR, our $\text{EventLens}_{\text{large}}$ demonstrates accuracy improvements of 12.3\%, 11.4\%, and 18.5\% across three sub-tasks of the test set with the $\text{OPT}_{\text{6.7b}}$ language model, indicating that task-specific models struggle to possess sufficient commonsense knowledge for Visual Commonsense Reasoning. 

Note that BLIP-2~\cite{blip2:li} itself has not been experimentally evaluated on the VCR task. Therefore, the data presented in Table~\ref{tab:main} here corresponds to experimental results obtained through fine-tuning on the VCR dataset using the pretrained BLIP-2 model (BLIP-2 ViT-g $\text{OPT}_{\text{6.7b}}$)." 

Further, from the perspective of pretrained Vision Language Transformers, our EventLens also achieves competitive performance against existing VLT-VCR models and outperforms UNITER-large by 5.4\%, 1.9\% and 5.7\% on three sub-tasks of the test set with less than half of the number of trainable parameters. Simultaneously, we observe that a stronger LLM leads to better performance. 

\begin{table}[t]
	\centering
	\resizebox{\columnwidth}{!}{%
		\begin{tabular}{l|cc|cc|cc}
			\hline
			\textbf{Model} 			& Q-A  & $\Delta$ & QA-R  &  $\Delta$ & Q-AR &  $\Delta$ \\ \hline \hline
			\textbf{EventLens-base}	 &80.6	&-		&79.9    	&-		    &64.7	    &- \\
			- w/o EA               		         &77.6   &3.0  &76.0       &3.9  	&59.9		&4.8  \\
			- w/o CLL         	                &78.1   &2.5   &78.2       &1.7      & 61.6	      &3.1 \\
			- w/o LoRA                        &78.3   &2.3   &78.1       &1.8       &60.5      &4.2\\ 
			Baseline(opt2.7b)             &73.8   &6.8    &72.3        &7.6      &54.4       & 10.3\\ \hline \hline
			\textbf{EventLens-large}  &82.0   & - &82.5  & -   &67.9   &-   \\
			- w/o EA               &78.7  &3.3   &77.5  & 5.0   &61.6   & 6.3  \\
			- w/o CLL        &79.2   &2.8  &80.3  &2.2    &64.1  &3.8    \\
			- w/o LoRA				&79.2  &2.8   &80.1 &2.4     &63.9   &6.0   \\ 
			Baseline(opt6.7b) &74.9   &7.1  &74.2  &8.3    &56.1    &11.8  \\ \hline
		\end{tabular}%
	}
	\caption{Ablation study of EventLens-base and EventLens-large on VCR validation set. Baseline(opt2.7b) and Baseline(opt6.7b) models are identical to BLIP-2 $\text{OPT}_{\text{2.7b}}$ and BLIP-2 $\text{OPT}_{\text{6.7b}}$ respectively.}
	\label{tab:ablation}
\end{table}

\subsection{Ablation Study}
We conduct ablation studies to evaluate the effectiveness of each module on VCR validation set, as shown in Table  \ref{tab:ablation}. Specifically, w/o EA means we remove the proposed Event-Aware Pretraining stage, w/o CLL means we remove the proposed Cross-modal Local Linking module during the Supervised Fine-tuning, and w/o LoRA means we remove VCR-LoRA adapters during supervised fine-tuning. From Table~\ref{tab:ablation}, we can observe that removing each module will result in an obvious  performance degradation, and removing Event-Aware Pretraining leads to most performance drop proving our proposed auxiliary task is essential for VCR task.

\subsection{Hyper-Parameter Analysis}
\begin{figure}[t]
	\centering
	\includegraphics[width=\columnwidth]{{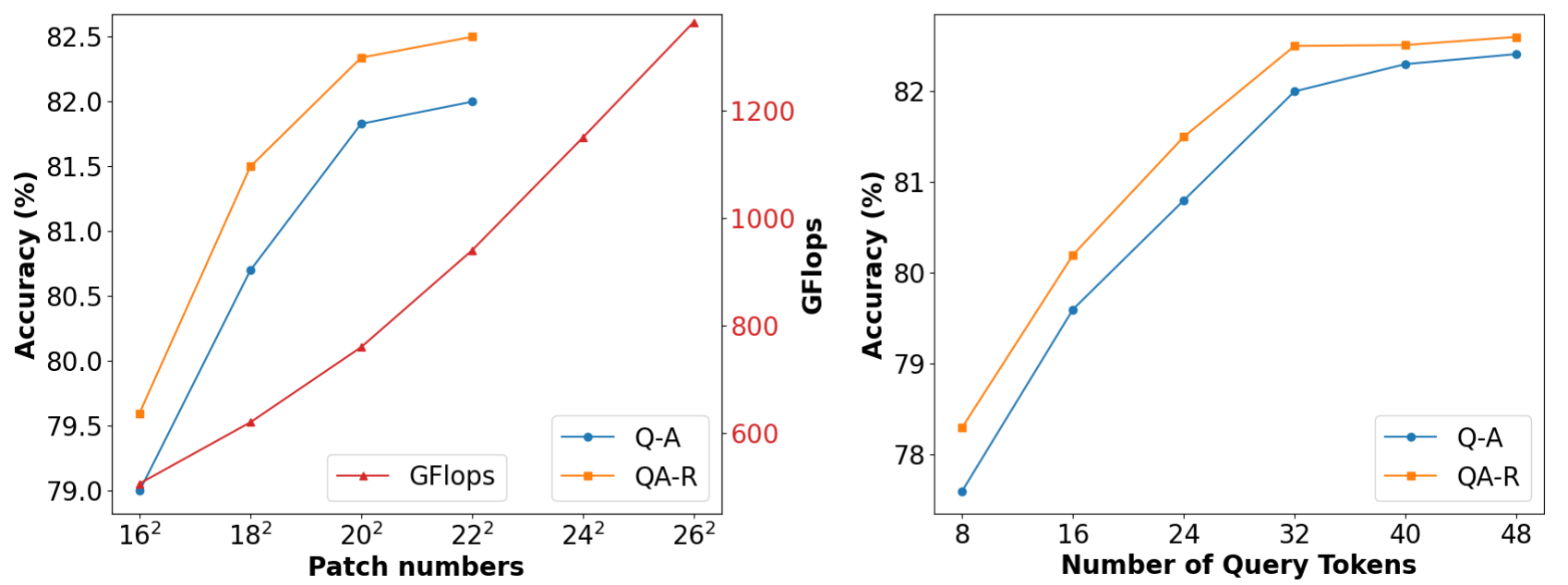}}
	\caption{Hyper-parameter analysis}
	\label{hyper}
\end{figure}

We conducted experimental analysis on key hyper-parameters of EventLens: the number of image patches and query tokens. Figure~\ref{hyper} illustrates our EventLens-large parameter analysis on the VCR validation set. The left subfigure shows the impact of varying input image patch numbers on VCR performance, exploring settings of $16^2$, $18^2$, $20^2$, and $22^2$ patches, all with a resolution of $14 \times 14$. Accuracy tends to saturate after the $20^2$ setting. The red line indicates the accelerating growth of floating-point operations for the Vision Encoder with increasing patches. Considering consumer-grade GPU constraints, we reported performance under the $22^2$ setting.

The right subfigure shows the influence of query tokens. We examined 8, 16, 24, 32, 40, and 48 query tokens, observing stabilization after the 32-token setting. Notably, the QA$\rightarrow$R subtask experiences a slight decline at 40 query tokens. This is attributed to longer text inputs in the QA$\rightarrow$R subtask, leading to truncation to maintain batch size and training efficiency. With more query tokens, the likelihood of truncated samples increases, impacting model performance. We chose to train our model with 32 query tokens.

\section{Conclusion}

In this paper, we propose EventLens, a Multimodal LLM architecture that leverages \textbf{Event}-Aware Pretraining and Cross-modal \textbf{L}inking and \textbf{En}hance\textbf{S} VCR. Specifically, we first introduce Event-Aware Pretraining task, requiring model to infer ongoing dynamic events and intentions of characters, to enhance model's commonsense knowledge reasoning abilities. Then we introduce a fine-grained Cross-modal Local Linking method to strengthen the model's comprehending of correlations between text tokens, local object features, and holistic complex scene. Finally, we use instruct-style prompts to narrow the gap between pretraining and VCR tasks, and task-specific adapters to better integrate LLM's inherent knowledge with new data paradigm. Experiments on VCR dataset demonstrate effectiveness of EventLens and obtains out-standing rank on leaderboard. For future work, we intend to explore VCR task from a causality perspective.

\bibliographystyle{named}
\bibliography{ijcai24_mmj}

\end{document}